\documentclass[10pt]{article}

\usepackage{tikz}
\usetikzlibrary{positioning, arrows.meta, calc}
\usepackage{float}
\usepackage[margin=1in]{geometry}
\usepackage{times}
\usepackage{graphicx}
\usepackage{amsmath}
\usepackage{amssymb}
\usepackage{booktabs}
\usepackage{caption}
\usepackage{enumitem}
\usepackage[numbers]{natbib}
\usepackage{hyperref}

\setlist[itemize]{leftmargin=*,itemsep=2pt,topsep=3pt}
\hypersetup{
    colorlinks=true,
    linkcolor=black,
    citecolor=black,
    urlcolor=black
}

\title{Do Transformer Temporal Heads and Post-Pooling Motion Gates Help CorrNet-based CSLR? An Empirical Study}

\author{
Lisi Wang$^{1}$,
Zhidong Xiao$^{2,*}$,
Jianjun Peng$^{3}$
}

\date{}

\begin{document}

\maketitle

\begin{center}
\small
$^{1}$University of New South Wales, Sydney, Australia\\
$^{2}$National Centre for Computer Animation, Faculty of Media, Science and Technology,\\
Bournemouth University, BH12 5BB, United Kingdom\\
$^{3}$School of Information Science and Engineering, Dalian Polytechnic University,\\
Dalian 116034, China\\[2mm]

\texttt{lisiwang.tech@gmail.com}\\
\texttt{zxiao@bournemouth.ac.uk},
\texttt{pengjj@dlpu.edu.cn}\\

$^{*}$Corresponding author
\end{center}

\begin{abstract}
CorrNet is a strong baseline for continuous sign language recognition (CSLR) because it models inter-frame correlations inside the visual encoding stage. In this paper, we study two natural extensions of a reproduced CorrNet system: replacing the BiLSTM temporal head with a Transformer encoder, and injecting motion cues after temporal pooling. We find that the Transformer head does not outperform the BiLSTM baseline, even with a training strategy adjusted for the Transformer, and the two heads have almost the same computational and runtime cost. For the second extension, we design a lightweight module called MotionGate. In our experiments, MotionGate consistently collapses to an identity-like mapping: the gate loses motion selectivity, and the injected residual becomes a weak, non-selective perturbation of the pooled features. These results suggest that explicit motion injection after CorrNet's correlation-based encoding is largely redundant, and that natural-looking architectural extensions in CSLR should be tested carefully instead of being assumed to help.
\end{abstract}

\section{Introduction}
\label{sec:introduction}

Continuous sign language recognition (CSLR) aims to recognize a sequence of glosses from an input signing video. Unlike isolated sign recognition, CSLR has to model both visual appearance and temporal dependencies over long video sequences, and training usually only uses sentence-level gloss annotations. Because of this, most modern CSLR systems follow a CTC-based pipeline: a visual backbone extracts frame-wise features, a temporal module models the resulting sequence, and CTC supervision aligns the predicted gloss sequence without frame-level labels.

A central challenge in CSLR is temporal modeling. Sign language contains rich motion patterns, such as hand trajectories, body movement, facial expressions, and transitions between signs. CorrNet handles this by modeling inter-frame correlations inside the visual encoder, which gives a strong representation for the downstream temporal module. This makes CorrNet a good basis for asking whether extra temporal or motion modules can still add useful information.

We study two natural extensions on a reproduced CorrNet system. The first replaces the original BiLSTM temporal module with a lightweight Transformer encoder. Self-attention can model global dependencies directly, so it is often seen as a stronger temporal model than recurrent networks. However, in the CorrNet pipeline, the sequence after downsampling is already short, and the backbone has already encoded local motion correlations. It is therefore not clear whether a Transformer head can bring extra gain over the recurrent baseline.

The second extension injects motion cues after temporal pooling. Early pooling may weaken fine-grained motion information, so post-pooling motion compensation looks like a reasonable fix. Based on this idea, we design a lightweight post-pooling motion gate, called MotionGate, which uses temporal feature differences to add residual motion information to the pooled sequence. We do not assume this module is helpful. Instead, we check whether the learned module actually contributes meaningful residual information in the reproduced CorrNet setting.

Our experiments show that neither extension improves recognition accuracy in
the reproduced CorrNet setting, while MotionGate converges to an identity-like
residual regime.

The main contributions of this paper are:
\begin{itemize}
    \item We conduct a controlled comparison between parameter-matched BiLSTM
    and Transformer temporal heads in a reproduced CorrNet-based CSLR pipeline.
    
    \item We introduce a lightweight post-pooling MotionGate and analyze its
    learned behavior using residual magnitude, injection strength, and
    gate--motion correlation diagnostics.
    
    \item On PHOENIX-2014, neither the Transformer head nor MotionGate improves
    recognition accuracy, and MotionGate converges to a weak, non-selective
    residual regime.
\end{itemize}

\section{Related Work}
\label{sec:related_work}

\subsection{Continuous Sign Language Recognition}

Continuous sign language recognition (CSLR) aims to recognize a sequence of glosses from an unsegmented signing video. Since frame-level gloss boundaries are usually unavailable, CSLR is commonly treated as a weakly supervised sequence learning problem. A common CSLR framework first extracts visual features with a CNN-based backbone, then uses a temporal module such as temporal convolution, BiLSTM, or Transformer to model the sequence, and finally applies a CTC classifier for gloss prediction \cite{graves2006connectionist,camgoz2017subunets,pu2019iterative,koller2020weakly,niu2020stochastic}.

Although this framework is effective, CSLR remains difficult because sign language videos contain complex hand, face, and body movements. The model needs to learn discriminative visual features, temporal dependencies, and the alignment between frames and gloss sequences at the same time. Earlier and contemporary CSLR systems have therefore explored hybrid CNN-LSTM-HMM modeling, iterative alignment, stochastic fine-grained gloss labeling, and multi-stream sequence learning to improve weakly supervised alignment and recognition \cite{koller2020weakly,pu2019iterative,niu2020stochastic}.

\subsection{CorrNet and VAC}

VAC studies the training difficulty of CTC-based CSLR. It argues that the visual feature extractor may not be trained well when supervision mainly comes from the final alignment module. To address this, VAC adds auxiliary visual supervision and prediction alignment: an auxiliary CTC loss on visual features, and an alignment loss between the visual branch and the main alignment branch \cite{min2021visual}. This idea is relevant to our baseline because the reproduced CorrNet-style training also uses an auxiliary convolutional CTC head and a distillation term, following knowledge distillation and self-mutual distillation ideas \cite{hinton2015distilling,hao2021self}. Other CSLR studies similarly strengthen CTC-based backbones through consistency constraints or self-emphasizing visual-temporal feature learning \cite{zuo2022c2slr,hu2023self}.

CorrNet improves the visual backbone by explicitly modeling cross-frame body trajectories. It computes correlation maps between the current frame and adjacent frames, so the model can capture local temporal movements such as hand and face trajectories \cite{hu2023continuous}. Compared with a plain CNN backbone, CorrNet adds motion-aware visual modeling before the final temporal module. In this work, we use CorrNet as the main reproduced baseline.

\subsection{Motion Cues}

Motion cues are important for sign language recognition because many signs are defined by movement direction, hand trajectory, speed, facial expression, and body dynamics. Existing CSLR methods model motion and spatial-temporal structure in different ways, including multi-cue spatial-temporal modeling, pose/keypoint streams, adaptive frame selection, temporal pooling, and correlation-based feature interaction \cite{zhou2020spatial,chen2022twostream,hu2023adabrowse,hu2022temporal,hu2023continuous}. This motivation is also related to general video recognition designs, where motion is modeled through optical-flow streams, 3D spatio-temporal convolution, efficient temporal shifting, or separate slow/fast temporal pathways \cite{simonyan2014two,carreira2017quo,lin2019tsm,feichtenhofer2019slowfast}.

CorrNet is one representative method that models motion-like information inside the visual backbone through correlations between adjacent frames \cite{hu2023continuous}. Temporal Lift Pooling studies temporal downsampling in CSLR and shows that the design of temporal pooling affects how well discriminative temporal information is preserved \cite{hu2022temporal}.

Our work studies a simple post-pooling motion compensation module. Instead of computing motion cues from raw frames or high-resolution visual features, we compute temporal differences from pooled feature sequences. This design is lightweight and easy to add after the visual backbone, but its usefulness needs to be tested against a strong CorrNet baseline.

\subsection{Transformer Temporal Modeling}

Transformers are widely used for sequence modeling because self-attention can model dependencies between any two time steps \cite{vaswani2017attention}. In sign language tasks, Transformer-based models have been used for both recognition and translation. For example, Sign Language Transformers jointly learn continuous sign language recognition and translation with a Transformer architecture and CTC loss \cite{camgoz2020sign}.

However, in CSLR, a Transformer temporal head does not always beat a BiLSTM. After visual encoding and temporal pooling, the sequence is often short, which limits the advantage of self-attention. Also, CSLR datasets are much smaller than common text datasets, and Transformer training can be sensitive to optimization and stabilization details \cite{liu2020understanding}. For these reasons, we compare Transformer and BiLSTM temporal heads under controlled settings.

\subsection{Speech Delta Features}

Speech recognition often uses dynamic acoustic features to describe local temporal changes. For example, MFCC features are often combined with delta and delta-delta features, which are first-order and second-order temporal derivatives. These features are simple but useful because they add short-term dynamic information to static frame-level representations \cite{furui1986speaker,davis1980comparison}.

Our motion compensation design is loosely inspired by this idea. We compute the difference between neighboring visual features and treat it as a motion-like signal. However, unlike speech delta features, our feature difference is computed after visual encoding and temporal pooling, not from low-level inputs. So we need to test whether such post-pooling differences still carry useful information for CSLR.

\section{Experimental Setup}
\label{sec:setup}

\subsection{Implementation Origin}

This study is built on the publicly available CorrNet implementation, which uses a ResNet-style visual backbone \cite{he2016deep} with a correlation module, and we use it as the reproduced CorrNet baseline \cite{hu2023continuous}. We do not claim the CorrNet backbone, the original data-processing pipeline, or the standard CTC training framework as our contributions.

Our modifications include:
\begin{itemize}
    \item a parameter-matched Transformer temporal head and its associated
    training adjustments;
    \item the post-pooling MotionGate module and its diagnostic measurements;
    \item a unified evaluation of WER, parameters, MACs, latency, and RTF.
\end{itemize}

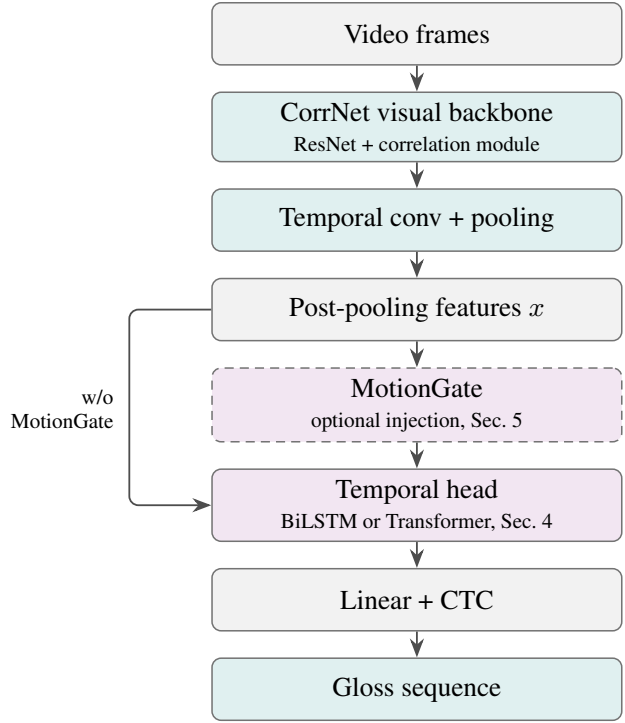
\begin{figure}[H]
\centering
\resizebox{0.55\linewidth}{!}{%
\begin{tikzpicture}[
  node distance=0.30cm,
  block/.style={
    rectangle,
    rounded corners=3pt,
    draw=black!55,
    line width=0.5pt,
    minimum width=4.75cm,
    minimum height=0.72cm,
    align=center,
    font=\small
  },
  arrow/.style={
    -{Stealth[length=2.2mm]},
    line width=0.6pt,
    draw=black!70
  }
]

\node[block, fill=gray!10] (video) {Video frames};

\node[block, fill=teal!12, below=of video] (backbone)
  {CorrNet visual backbone\\[-1pt]
  {\scriptsize ResNet + correlation module}};

\node[block, fill=teal!12, below=of backbone] (tconv)
  {Temporal conv + pooling};

\node[block, fill=gray!10, below=of tconv] (feat)
  {Post-pooling features $x$};

\node[block, fill=violet!10, densely dashed, below=of feat] (mg)
  {MotionGate\\[-1pt]
  {\scriptsize optional injection, Sec.~\ref{sec:motion_study}}};

\node[block, fill=violet!10, below=of mg] (head)
  {Temporal head\\[-1pt]
  {\scriptsize BiLSTM or Transformer, Sec.~\ref{sec:temporal_study}}};

\node[block, fill=gray!10, below=of head] (ctc)
  {Linear + CTC};

\node[block, fill=teal!12, below=of ctc] (out)
  {Gloss sequence};

\draw[arrow] (video)    -- (backbone);
\draw[arrow] (backbone) -- (tconv);
\draw[arrow] (tconv)    -- (feat);
\draw[arrow] (feat)     -- (mg);
\draw[arrow] (mg)       -- (head);
\draw[arrow] (head)     -- (ctc);
\draw[arrow] (ctc)      -- (out);

\coordinate (bp) at ($(feat.west)+(-0.95,0)$);

\draw[arrow, rounded corners=3pt]
  (feat.west) -- (bp) |- (head.west);

\node[font=\scriptsize, anchor=east, align=right]
  at ($(bp)+(-0.06,-1.15)$)
  {w/o\\MotionGate};

\path ($(feat.east)+(0.75,0)$);
\end{tikzpicture}%
}
\caption{Overall pipeline of the reproduced CorrNet-based CSLR system.
Video frames are encoded by the CorrNet visual backbone, downsampled by
temporal convolution and pooling, and then recognized by a temporal head
under CTC supervision. MotionGate is an optional post-pooling module studied
in Sec.~\ref{sec:motion_study}, while the BiLSTM and Transformer temporal
heads are compared in Sec.~\ref{sec:temporal_study}. }
\label{fig:pipeline}
\end{figure}

\subsection{Dataset and Evaluation Protocol}

We conduct all experiments on PHOENIX-2014 \cite{koller2015continuous}. Figure~\ref{fig:pipeline} shows the overall pipeline and where our two studied components are inserted. The model follows the standard CTC-based CSLR pipeline: a visual backbone extracts frame-wise representations from the input video, a temporal module models the feature sequence, and CTC supervision is applied at the gloss level without frame-level annotations.

Recognition accuracy is measured by gloss-level word error rate (WER) on the development and test sets. We follow the standard PHOENIX-2014 WER convention, but compute WER with a Python implementation of edit distance instead of the SCTK \texttt{sclite} toolkit. All compared models are evaluated with the same decoding and scoring pipeline, so the results are directly comparable within this study. However, since the scoring implementation differs from the official SCTK-based protocol used in some prior work, the absolute WER values should mainly be used for internal comparison, not for direct comparison with published numbers.

Unless otherwise stated, each reported configuration corresponds to a single training run with a fixed random seed, and we do not report variance across seeds. We discuss what this means for interpretation in Section~\ref{sec:limitations}.

\subsection{Efficiency Metrics}

Besides recognition accuracy, we report efficiency metrics: number of trainable parameters, multiply-accumulate operations (MACs), model-only inference latency, and real-time factor (RTF). We report MACs instead of strict FLOPs because the profiling tool counts multiply-accumulate operations. Latency and RTF are measured on the full PHOENIX-2014 test split with batch size 1, excluding beam-search decoding. Runtime numbers depend on hardware, so all latency and RTF results are measured on the same platform with an NVIDIA RTX 5090 GPU and an AMD EPYC 9J14 CPU.

\subsection{Training Configuration Note}

The BiLSTM results in the Temporal Modeling Study and the Motion Compensation Study use different training configurations. The former uses the Transformer-specific training strategy for a fair temporal-head comparison, and the latter uses the standard 1:1 supervision configuration as the reference for MotionGate. So comparisons are only made within each study, not across the two studies. The standard BiLSTM and MotionGate configurations use Adam \cite{kingma2015adam}, while the Transformer-specific temporal-head comparison uses AdamW \cite{loshchilov2019decoupled} for both the BiLSTM and Transformer heads.

\section{Temporal Modeling Study}
\label{sec:temporal_study}

\subsection{Motivation and Protocol}

The original CorrNet model uses a BiLSTM temporal module after the visual backbone \cite{hochreiter1997long,graves2005framewise}. To check whether global self-attention \cite{vaswani2017attention} can further improve CorrNet-based CSLR, we replace the BiLSTM with a lightweight Transformer encoder, while keeping the visual backbone, CTC supervision, decoding, and evaluation pipeline unchanged. This isolates the effect of the temporal modeling component under the same overall CorrNet framework.

We first trained the Transformer variant with the same loss configuration as the reproduced CorrNet baseline, where the visual head and the temporal sequence head have equal weights. Under this configuration, the Transformer head trains less stably and gets worse accuracy. We therefore use a Transformer-specific training configuration with stronger supervision on the temporal sequence branch in early training. Specifically, we use AdamW \cite{loshchilov2019decoupled}, increase the relative weight of the sequence loss, use adaptive distillation adapted from knowledge distillation \cite{hinton2015distilling} and the distillation strategy in the original CorrNet framework~\cite{hu2023continuous}, and reduce the distillation temperature. For a fair comparison, the BiLSTM result in this study is trained with the same Transformer-specific strategy.

\paragraph{Transformer architecture details.}
The Transformer temporal head is a pre-norm Transformer encoder with 2 layers,
8 attention heads, \(d_{\text{model}} = 1024\), and
\(d_{\text{ff}} = 1024\). Here, \(d_{\text{model}}\) is the hidden dimension of the Transformer features,
while \(d_{\text{ff}}\) is the inner dimension of the feed-forward sublayer.
We set \(d_{\text{ff}} = d_{\text{model}}\) to keep the Transformer head
parameter-matched with the BiLSTM head. A linear input projection is used when the incoming
temporal feature dimension differs from \(d_{\text{model}}\). We use sinusoidal
positional encoding with a maximum length of 5000, dropout of 0.1, and GELU
activations. The temporal-head parameter count, 12.6054M, closely
matches that of the BiLSTM head, 12.5993M, so the comparison in Table~\ref{tab:temporal_head_comparison} mainly
reflects the temporal modeling mechanism rather than model capacity.

\subsection{Results and Analysis}

Table~\ref{tab:temporal_head_comparison} compares the BiLSTM and Transformer temporal heads under the reproduced CorrNet setting. Both entries are trained with the Transformer-specific strategy, including AdamW, adjusted head weights, and adaptive distillation. Even though the adjusted training improves the Transformer variant, it still does not outperform the BiLSTM. The BiLSTM reaches 21.39\% dev WER and 21.60\% test WER, while the Transformer gets 22.53\% and 23.55\%.

The efficiency results show that this gap is not caused by a difference in model scale or computational budget. The two heads have almost the same temporal parameter counts: 12.5993M for BiLSTM and 12.6054M for Transformer. Their full-model MACs, model-only latency, and RTF are also very close. In other words, the Transformer head gives worse accuracy at nearly the same computational and runtime cost.

A likely reason is that the temporal sequence after downsampling is already short, around 40 time steps in our setting. At this scale, self-attention has little room to exploit long-range dependencies, and its pairwise interaction mechanism does not give a clear return. The moderate temporal resolution after two pooling stages and the limited size of PHOENIX-2014 may further reduce the benefit of training an attention-based head from scratch. We therefore keep the BiLSTM temporal head in the final model.

\begin{table}[t]
\centering
\caption{Temporal-head comparison under the Transformer-specific training strategy. Each entry is a single training run (see Section~\ref{sec:limitations}).}
\label{tab:temporal_head_comparison}
\resizebox{\linewidth}{!}{%
\begin{tabular}{@{}lcccccc@{}}
\toprule
\textbf{Temporal Head} & \textbf{Dev WER (\%)} & \textbf{Test WER (\%)} & \textbf{Head Params (M)} & \textbf{MACs (G)} & \textbf{Latency (ms/sample)} & \textbf{RTF} \\
\midrule
BiLSTM      & \textbf{21.39} & \textbf{21.60} & \textbf{12.60} & 268.90          & \textbf{29.93} & \textbf{0.0053} \\
Transformer & 22.53          & 23.55          & 12.61          & \textbf{268.63} & 30.74          & 0.0054 \\
\bottomrule
\end{tabular}%
}
\end{table}

\section{Motion Compensation Study}
\label{sec:motion_study}

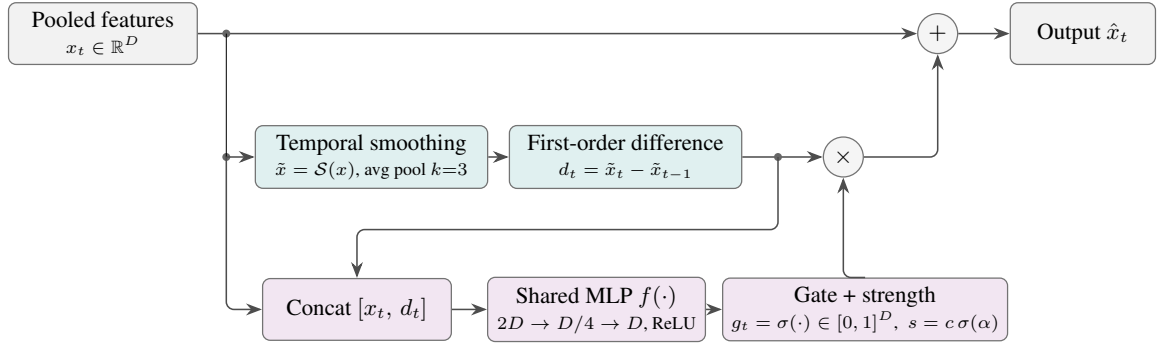
\begin{figure}[H]
\centering
\resizebox{0.92\linewidth}{!}{%
\begin{tikzpicture}[
  block/.style={
    rectangle,
    rounded corners=3pt,
    draw=black!55,
    line width=0.5pt,
    minimum height=0.85cm,
    align=center,
    font=\small
  },
  op/.style={
    circle,
    draw=black!55,
    line width=0.5pt,
    fill=gray!8,
    inner sep=1pt,
    minimum size=0.55cm,
    font=\small
  },
  arrow/.style={
    -{Stealth[length=2.2mm]},
    line width=0.6pt,
    draw=black!70
  },
  jdot/.style={
    circle,
    fill=black!60,
    inner sep=1.1pt
  }
]

\node[block, fill=gray!10, minimum width=2.6cm] (x) at (1.4,0)
  {Pooled features\\[-1pt] {\scriptsize $x_t \in \mathbb{R}^{D}$}};

\node[op] (add) at (12.9,0) {$+$};

\node[block, fill=gray!10, minimum width=2.0cm] (xhat) at (14.9,0)
  {Output $\hat{x}_t$};

\draw[arrow] (x.east) -- (add.west);
\draw[arrow] (add.east) -- (xhat.west);

\node[block, fill=teal!12, minimum width=3.2cm] (smooth) at (5.1,-1.7)
  {Temporal smoothing\\[-1pt]
  {\scriptsize $\tilde{x} = \mathcal{S}(x)$, avg pool $k{=}3$}};

\node[block, fill=teal!12, minimum width=3.2cm] (diff) at (8.6,-1.7)
  {First-order difference\\[-1pt]
  {\scriptsize $d_t = \tilde{x}_t - \tilde{x}_{t-1}$}};

\node[op] (mult) at (11.6,-1.7) {$\times$};

\draw[arrow] (smooth.east) -- (diff.west);
\draw[arrow] (diff.east) -- (mult.west);
\draw[arrow, rounded corners=3pt] (mult.east) -| (add.south);

\node[block, fill=violet!10, minimum width=2.6cm] (cat) at (4.9,-3.8)
  {Concat $[x_t,\, d_t]$};

\node[block, fill=violet!10, minimum width=3.0cm] (mlp) at (8.2,-3.8)
  {Shared MLP $f(\cdot)$\\[-1pt]
  {\scriptsize $2D \to D/4 \to D$, ReLU}};

\node[block, fill=violet!10, minimum width=3.6cm] (gate) at (11.9,-3.8)
  {Gate + strength\\[-1pt]
  {\scriptsize $g_t = \sigma(\cdot) \in [0,1]^{D}$,\; $s = c\,\sigma(\alpha)$}};

\draw[arrow] (cat.east) -- (mlp.west);
\draw[arrow] (mlp.east) -- (gate.west);
\draw[arrow, rounded corners=3pt] (gate.north) -| (mult.south);

\node[jdot] (j1) at (3.1,0) {};
\node[jdot] (j2) at (3.1,-1.7) {};

\draw (j1.center) -- (j2.center);
\draw[arrow] (j2.center) -- (smooth.west);
\draw[arrow, rounded corners=3pt] (j2.center) |- (cat.west);

\node[jdot] (j3) at (10.7,-1.7) {};

\draw[arrow, rounded corners=3pt]
  (j3.center) -- (10.7,-2.7) -- (4.9,-2.7) -- (cat.north);

\end{tikzpicture}%
}
\caption{Internal structure of MotionGate.}
\label{fig:motiongate}
\end{figure}

\subsection{Post-Pooling Motion Compensation}

\paragraph{Motivation.}
CSLR relies heavily on temporal motion cues, such as hand trajectories, movement direction, velocity, and body displacement, not only static frame-level appearance. In CorrNet, the visual features are followed by temporal pooling, which reduces temporal resolution. A natural concern is that early temporal pooling may weaken fine-grained or high-frequency motion cues. A natural hypothesis is that explicit temporal differences may re-emphasize motion-related variations that remain in the pooled representation.

We treat this as a hypothesis to test, not as an assumed contribution. The goal is to check whether post-pooling temporal differences still give useful and generalizable information once the CorrNet backbone has already encoded visual-temporal correlations.

\paragraph{MotionGate.}
Based on this hypothesis, we design a lightweight module called MotionGate. Figure~\ref{fig:motiongate} illustrates the module structure. Given a pooled feature sequence
\[
x = \{x_t\}_{t=1}^{T}, \qquad x_t \in \mathbb{R}^{D},
\]
we first smooth the pooled features and then compute a first-order temporal difference:
\[
\tilde{x} = \mathcal{S}(x),
\qquad
d_t = \tilde{x}_t - \tilde{x}_{t-1},
\]
where $\mathcal{S}(\cdot)$ is local temporal smoothing (average pooling with kernel size 3) used to suppress frame-to-frame high-frequency noise. The motion signal is then filtered by a channel-wise one-sided gate:
\[
g_t = \sigma\left(f([x_t, d_t])\right),
\qquad
g_t \in [0,1]^D,
\]
where $f(\cdot)$ is a small MLP and $\sigma(\cdot)$ is the sigmoid function. The final output uses additive residual injection:
\[
\hat{x}_t = x_t + s \cdot g_t \odot d_t,
\]
where $s$ is a learnable scalar constrained by a predefined magnitude cap, and $\odot$ is channel-wise multiplication. So $s \cdot g_t$ controls how much motion information is injected into each channel.

The scalar strength $s$ is implemented with a sigmoid-parameterized logit and a fixed upper bound:
\[
s = c \cdot \sigma(\alpha),
\]
where $c$ is the magnitude cap and $\alpha$ is a learnable parameter. This makes sure the residual branch cannot amplify the temporal-difference signal without limit. The module is initialized so that the initial output stays close to the original pooled feature:
\[
\hat{x}_t \approx x_t.
\]
This avoids disturbing the pretrained visual backbone at the start of training.

We also tried second-order temporal differences and unconstrained residual injection without a magnitude cap. These early variants were prone to overfitting and did not improve recognition, likely because second-order differences amplify high-frequency noise, and unconstrained injection lets the residual branch dominate the pooled representation. These variants involve acceleration terms and were not run as clean isolated comparisons, so we only use them as design motivation, not as standalone experimental evidence. The final MotionGate design therefore uses smoothed first-order temporal difference with magnitude-capped additive residual injection.

\paragraph{Ruling Out Insufficient Optimization.}
One alternative explanation is that MotionGate collapses not because the post-pooling temporal-difference signal is redundant, but because the motion branch gets too little gradient or supervision under the original training setup. This concern is reasonable: the motion module sits near the convolutional stage, and the original setting uses a high distillation temperature and a relatively low convolutional-head supervision weight, both of which may weaken the learning signal for the motion branch.

To rule this out, we train MotionGate under a configuration that deliberately favors the motion branch. We use the standard 1:1 visual/temporal supervision configuration during the whole training, so the convolutional-stage gradient is not down-weighted, and we give the MotionGate parameters a separate higher learning rate of $1 \times 10^{-3}$. This setting maximizes the chance that the motion branch can learn, if the post-pooling temporal-difference signal is actually useful.

\begin{figure}[H]
    \centering
    \includegraphics[width=\linewidth]{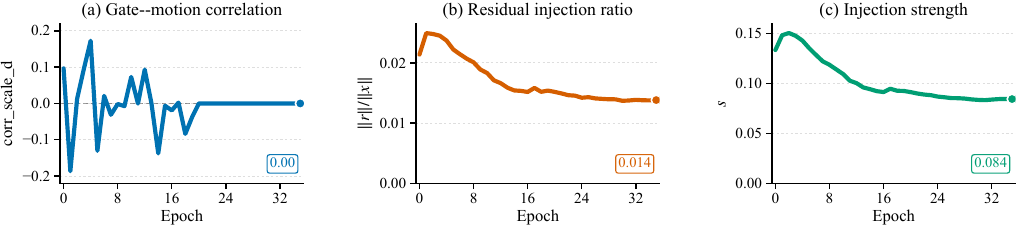}
    \caption{MotionGate collapse diagnostics under the matched 1:1 supervision setting.}
    \label{fig:motiongate_collapse}
\end{figure}

Even under this favorable configuration, the gate still collapses. As shown in
Figure~\ref{fig:motiongate_collapse}, \texttt{corr\_scale\_d} drops toward
zero. This metric measures the correlation between the temporal profiles of
\(|a_t|\) and \(|d_t|\) after channel averaging, where \(a_t\) denotes the
effective residual-injection scale produced by the gate and scalar strength
(the variable \texttt{scale} in the implementation). Values near zero indicate
that the gate is not selectively responding to stronger temporal-difference
positions. At the same time, the residual-to-feature norm ratio
\[
\texttt{res\_x\_norm\_ratio}
=
\frac{\|a \odot d\|_2}{\|x\|_2}
\]
stays very small, around 0.014 in late training, and the learned injection
strength remains low. These diagnostics suggest that MotionGate does not learn
a selective motion-compensation mechanism; the injected residual instead
becomes a weak, non-selective perturbation of the pooled feature. We call this
end state an \emph{identity-like mapping}: the module output stays numerically
close to, but not exactly equal to, the input feature.

Because the gate loses selectivity even when gradient strength and supervision are made favorable, the collapse cannot be explained only by insufficient optimization. It suggests that the post-pooling temporal-difference signal itself cannot be stably exploited by the model.

\paragraph{Recognition Performance.}
We further compare the BiLSTM baseline and BiLSTM with MotionGate under the same standard 1:1 supervision configuration. As shown in Table~\ref{tab:motiongate_bilstm}, adding MotionGate does not improve recognition accuracy. The BiLSTM baseline gets 20.43\% dev WER and 21.03\% test WER, while BiLSTM + MotionGate gets 21.23\% and 21.46\%. The difference is small and should not be over-read as a real degradation, especially because each configuration is trained only once and we have no seed-variance estimates (Section~\ref{sec:limitations}). The more reliable conclusion is that MotionGate gives no recognition gain under this controlled setting.

MotionGate also adds parameters and computational overhead. The parameter count grows from 33.37M to 34.16M, and latency grows from 29.50 ms/sample to 31.13 ms/sample. Since the module does not reduce WER, this extra complexity is not justified.

\begin{table}[t]
\centering
\caption{MotionGate comparison under the standard 1:1 supervision configuration. Each entry is a single training run (see Section~\ref{sec:limitations}).}
\label{tab:motiongate_bilstm}
\resizebox{\linewidth}{!}{%
\begin{tabular}{@{}lcccccc@{}}
\toprule
\textbf{Method} & \textbf{Dev WER (\%)} & \textbf{Test WER (\%)} & \textbf{Params (M)} & \textbf{MACs (G)} & \textbf{Latency (ms/sample)} & \textbf{RTF} \\
\midrule
BiLSTM baseline     & \textbf{20.43} & \textbf{21.03} & \textbf{33.37} & \textbf{268.91} & \textbf{29.50} & \textbf{0.00518} \\
BiLSTM + MotionGate & 21.23          & 21.46          & 34.16          & 268.96          & 31.13          & 0.00547 \\
\bottomrule
\end{tabular}%
}
\end{table}

\paragraph{Interpretation.}
Under favorable optimization conditions, MotionGate consistently collapses to the identity-like mapping described above: the gate loses motion selectivity, and the residual branch remains only a weak, non-selective perturbation. This is shown by \texttt{corr\_scale\_d} approaching zero, together with a small but nonzero \texttt{res\_x\_norm\_ratio} and a low learned injection strength. At the same time, recognition accuracy does not improve under the matched 1:1 supervision setting. Since this behavior persists even with stronger gradient and supervision for the motion branch, it cannot be attributed to insufficient optimization.

Instead, the result indicates that the post-pooling temporal-difference signal is largely redundant once CorrNet has already encoded visual-temporal correlations. Explicit temporal differences do not give a stable extra cue for improving generalization after temporal pooling. The model therefore fails to learn a selective motion-compensation mechanism and reduces MotionGate to a negligible non-selective residual perturbation.

\section{Discussion}
\label{sec:discussion}

Our experiments show that two natural extensions to CorrNet, replacing the BiLSTM temporal head with a Transformer encoder and injecting post-pooling motion compensation through MotionGate, do not improve recognition accuracy under the reproduced PHOENIX-2014 setting. This does not reject Transformers or motion modeling for CSLR in general. It shows that their effectiveness depends strongly on where they are inserted and how much usable temporal information is left at that stage.

\paragraph{Why the Transformer head does not improve over BiLSTM.}
A Transformer encoder is attractive for sequence modeling because self-attention can directly model long-range dependencies. However, in our setting the temporal sequence has already been shortened by the convolutional backbone and temporal pooling to roughly 40 time steps. At this scale, the long-range advantage of self-attention is weak, and its quadratic pairwise interaction does not turn into a clear recognition gain. The Transformer gives a more flexible temporal operator, but the post-pooling sequence does not seem to need this extra flexibility.

Another issue is inductive bias. BiLSTMs impose a strong sequential prior by processing the feature sequence recurrently in both directions. This fits CSLR well, where local order, monotonic temporal progression, and gradual motion transitions matter. A Transformer instead relies more on data and positional encoding to learn such structure. On a relatively small CSLR dataset, and after the backbone has already compressed the temporal signal, this weaker sequential prior may make the Transformer less data-efficient than the BiLSTM. This is likely why even Transformer-specific training adjustments and stronger temporal supervision do not lead to a clear improvement.

\paragraph{Why post-pooling MotionGate becomes redundant.}
MotionGate comes from a natural concern: temporal pooling may weaken fine-grained motion cues, and explicit temporal differences might compensate for this. However, the diagnostics show that the gate loses motion selectivity and turns into a weak, non-selective residual injection. There are several possible reasons.

First, from an information view, the temporal difference signal is a deterministic function of the pooled feature sequence:
\[
d_t = \tilde{x}_t - \tilde{x}_{t-1}.
\]
So the pair $(x, d)$ contains no information that is absent from $x$ itself. Adding $d$ may change the parameterization or the optimization path, but it cannot recover information already removed by pooling.

Second, CorrNet is not a purely frame-wise visual encoder. Its correlation mechanism explicitly models temporal relations and frame-to-frame visual correspondences before temporal pooling. Motion-related information has therefore already been encoded by the backbone before MotionGate is applied. Injecting first-order differences after pooling is more likely to duplicate this representation than to complement it.

Third, the motion information removed by pooling may simply matter little for gloss-level recognition. Temporal pooling mainly suppresses high-frequency variations, and PHOENIX-2014 recognition is evaluated at the sequence-label level with CTC supervision. The objective rewards stable discriminative patterns for gloss prediction, not precise reconstruction of every fine-grained motion fluctuation. So the residual motion signal after pooling may contribute little to WER.

For these reasons, MotionGate collapses even under favorable optimization conditions. The module gets 1:1 visual/temporal supervision and a higher learning rate for its parameters, yet it still fails to learn a selective compensation mechanism. The residual stays small and non-selective because the explicit post-pooling difference largely overlaps with the information CorrNet has already encoded.

The two studies show a similar pattern: when the backbone already does strong temporal correlation modeling, simple temporal extensions that look natural on their own can become redundant after temporal compression. The Transformer head adds a more expressive sequence model, but the post-pooling sequence is short and already well structured. MotionGate adds explicit temporal differences, but these are deterministic functions of already pooled features and overlap with CorrNet's correlation modeling.

This suggests that CSLR improvements should not be expected from adding more temporal machinery alone. A temporal module is more likely to help when it can access information the backbone has not encoded yet, or when it operates before irreversible temporal compression. After a strong correlation-based backbone and temporal pooling, extra temporal or motion modules may mainly add complexity without improving generalization.

\section{Limitations}
\label{sec:limitations}

We note three limitations of this study.

\paragraph{Single-run results.}
Each configuration is evaluated using one fixed-seed training run, so small
WER differences may reflect seed-level variation. We therefore interpret the
MotionGate result as an absence of improvement rather than evidence of a
significant degradation.

\paragraph{Non-standard WER scoring implementation.}
WER is computed with a Python implementation of edit distance, not the official SCTK \texttt{sclite} toolkit used in much of the CSLR literature. All models in this paper are scored with the same pipeline, so within-paper comparisons remain valid, but the absolute WER values should not be directly compared with published numbers obtained under the \texttt{sclite} protocol.

\paragraph{Single dataset and single backbone.}
All experiments use PHOENIX-2014 and a reproduced CorrNet backbone. The observed redundancy of post-pooling motion injection and the lack of benefit from a Transformer temporal head are conclusions about this specific setting. They may not transfer to larger or more diverse CSLR datasets (e.g., PHOENIX-2014T or CSL-Daily), to backbones without explicit correlation-based temporal modeling, or to pipelines with different pooling configurations, where the effective sequence length and the remaining motion information may be quite different. Latency and RTF numbers are also specific to the hardware platform described in Section~\ref{sec:setup}.

\section{Conclusion}
\label{sec:conclusion}

We conducted a controlled study of temporal modeling extensions for CorrNet-based continuous sign language recognition on PHOENIX-2014. We examined two natural modifications: replacing the BiLSTM temporal head with a Transformer encoder, and adding post-pooling motion compensation through MotionGate. Under matched training and evaluation settings, neither extension improves over the BiLSTM baseline.

The Transformer head gives no clear recognition benefit despite Transformer-specific training adjustments, suggesting that self-attention offers limited advantage once CorrNet has shortened the temporal sequence and encoded visual-temporal correlations. MotionGate also fails to improve accuracy. Its diagnostics show that it collapses to an identity-like mapping, losing motion selectivity and degenerating into a weak, non-selective residual injection, which indicates that post-pooling temporal differences are largely redundant in this setting.

Based on these findings, we keep the BiLSTM baseline as the final temporal model. More broadly, the results show that natural-looking temporal or motion-based extensions need careful validation when applied after a strong correlation-based backbone. Future work may study whether Transformers become more effective under different pooling configurations, or whether motion injection is more useful before temporal compression or at earlier visual feature stages.

\bibliographystyle{plainnat}
\bibliography{references}

\end{document}